\documentclass[12pt,letterpaper]{article}

\usepackage[pagenumbers]{cvpr} % To force page numbers, e.g. for an arXiv version

\usepackage{amsmath}
\usepackage{amssymb}

\usepackage{tabulary}
\usepackage{colortbl}
\definecolor{Gray}{gray}{0.9}
\definecolor{citecolor}{HTML}{2980b9}
\definecolor{linkcolor}{HTML}{c0392b}
\definecolor{deemph}{gray}{0.6}

\definecolor{baselinecolor}{gray}{.9}

\usepackage{longtable}
\usepackage{bm}

\usepackage{algorithm}
\usepackage{algorithmic}

\definecolor{cvprblue}{rgb}{0.21,0.49,0.74}
\usepackage[pagebackref,breaklinks,colorlinks,allcolors=cvprblue]{hyperref}

\def\paperID{*****} % *** Enter the Paper ID here
\def\confName{CVPR}
\def\confYear{2025}

\title{Powerful Design of Small Vision Transformer on CIFAR10}
\author{Gent Wu\\
Surrey Institute for People-Centred AI (PAI)\\
 Guildford GU2 7XH, Surrey, United Kingdom\\
{\tt\small clouderow@gmail.com}
}
\begin{document}
\maketitle
\begin{abstract}

Vision Transformers (ViTs) have demonstrated remarkable success on large-scale datasets, but their performance on smaller datasets often falls short of convolutional neural networks (CNNs). This paper explores the design and optimization of Tiny ViTs for small datasets, using CIFAR-10 as a benchmark. We systematically evaluate the impact of data augmentation, patch token initialization, low-rank compression, and multi-class token strategies on model performance. Our experiments reveal that low-rank compression of queries in Multi-Head Latent Attention (MLA) incurs minimal performance loss, indicating redundancy in ViTs. Additionally, introducing multiple CLS tokens improves global representation capacity, boosting accuracy. These findings provide a comprehensive framework for optimizing Tiny ViTs, offering practical insights for efficient and effective designs. Code is available at \url{https://github.com/erow/PoorViTs}

\end{abstract}    
Vision Transformers (ViTs) have achieved state-of-the-art results on several benchmarks. However, their performance gains heavily depend on large-scale training, either through supervised learning~\citep{dosovitskiy2020image} or self-supervised learning~\citep{atito2021sit}. The high computational cost associated with ViTs has hindered architectural innovation. Moreover, smaller ViTs often exhibit inferior effectiveness and efficiency compared to their convolutional counterparts. For instance, a Tiny ViT achieves only 80\% accuracy on CIFAR-10, as reported by \citeauthor{Cydia2018}. While some studies have attempted to bridge the performance gap for small ViTs on small datasets, their solutions often fail to fully leverage the power of Transformers~\citep{vaswani2017attention} by relying on convolutional techniques~\citep{shao2022transformers,ZHANGconvvit}. For example, \citeauthor{ZHANGconvvit} improved ViT performance by incorporating depth-wise convolutions, achieving 96.41\% accuracy. Other works have explored self-supervised learning to enhance feature representation~\citep{liu2021efficient,gani2022train}. Our goal is to establish an efficient yet effective framework for training small ViTs on small datasets, facilitating research, particularly for beginners.

The CIFAR-10 dataset, introduced by \citeauthor{krizhevsky2009learning}, has become a widely adopted benchmark for evaluating image classification models in machine learning. It comprises 60,000 32x32 color images distributed across 10 classes, with 6,000 images per class. The dataset is divided into 50,000 training images and 10,000 test images, providing a balanced platform for rapid experimentation and model validation. Despite its simplicity, CIFAR-10 remains a critical tool for the rapid validation of new algorithms. Recent advancements in training methods have significantly improved the speed and accuracy of neural networks on CIFAR-10. For instance, \citeauthor{jordan202494} introduced training methods that achieve 94\% accuracy in just 3.29 seconds on a single NVIDIA A100 GPU. However, recent progress in efficient training on CIFAR-10 has primarily focused on convolutional neural networks (CNNs). 

In this report, we investigate the architectural design of a Tiny ViT~\citep{gani2022train} to achieve competitive performance compared to CNN counterparts (93.58\% on CIFAR10). Our findings reveal two key insights: 1) applying low-rank compression to queries does not degrade performance, and 2) reducing the dimensionality of patch tokens while maintaining the dimensionality of the CLS token does not diminish performance. These results suggest the presence of redundant information in ViTs, which can be exploited to improve efficiency without sacrificing accuracy.
\section{Vision Transformer}\label{sec:method}

\begin{figure}
    \centering
    \includegraphics[width=0.75\linewidth]{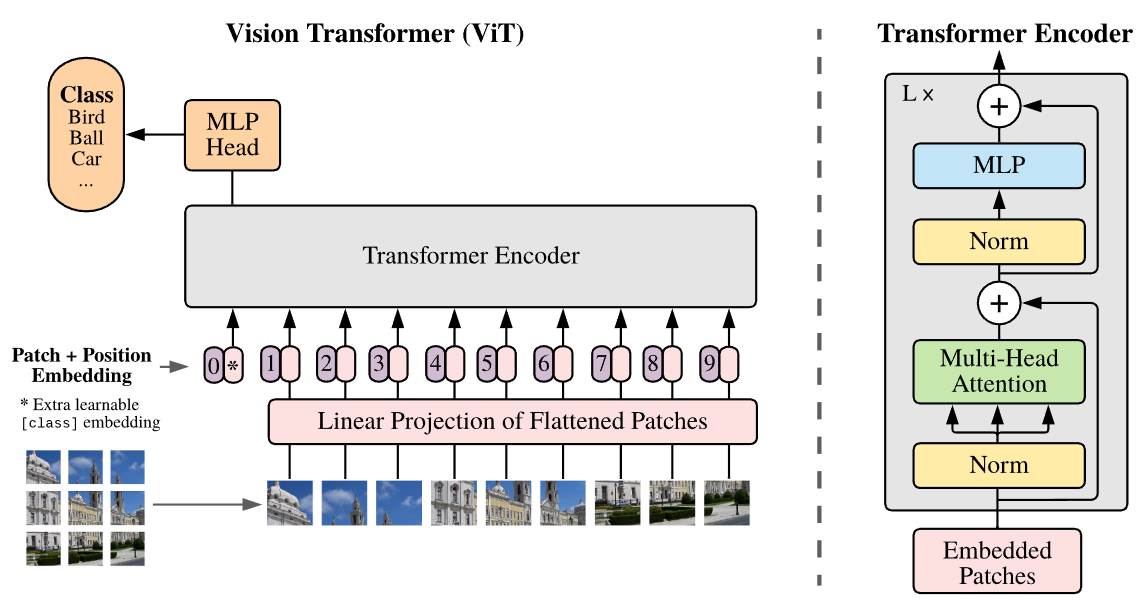}
    \caption{Vision transformer overview from \cite{dosovitskiy2020image}. It involves (1) \textbf{Tokenization}: The image is divided into patches of a predetermined size. Each patch is then linearly embedded, and position embeddings are incorporated to maintain spatial information. The sequence of vectors, which represents the embedded patches along with their positional data. (2) \textbf{Token transformation}: tokens are subsequently input into a conventional Transformer encoder. (3) \textbf{Task projection}: For the purpose of classification, a common method is employed, which involves appending an additional learnable "classification token" to the sequence of vectors. This token is trained to aggregate the information from all patches and serves as the basis for the classification decision.
}
    \label{fig:vit}
\end{figure}

Vision Transformer (ViT), as illustrated in \cref{fig:vit}, was first proposed by \citeauthor{dosovitskiy2020image}, drawing inspiration from the success of Transformers~\citep{vaswani2017attention} in natural language processing. Transformers utilize an attention mechanism to model the relationships between tokens effectively. In ViTs, images are split into small patches and processed to extract high-level features. Specifically, a ViT has the following key procedures: image tokenization, token transformation, and task projection. 

\subsection{Patch Token = Patch Embedding + Positional Embedding} 
Tokenization generates patch tokens through patch embedding and positional embedding. Specifically, images are split into several patches, and each patch token represents the semantic information at a specific position. Initially, patch tokens only contain low-level pixel information. To differentiate patches with similar colors or textures, positional embeddings are essential. For instance, without positional embeddings, two patches from a white T-shirt and a white bag would be indistinguishable and fail to represent different objects (T-shirt and bag here). The tokenization can be formalized as: input image $\textbf{x}_0 \in \mathbb{R}^{3 \times H \times W}$ is rearranged into a patch sequence $\textbf{x} \in \mathbb{R}^{L \times (3 \times P^2)}$, where $P$ is the patch size and $L = HW / P^2$ is the number of patches. A patch embedding layer projects each patch into a higher-dimensional space to obtain the patch embeddings $\textbf{x}_e \in \mathbb{R}^{L \times C}$, where $C$ is the embedding dimension. Position embeddings $\mathbf{x}_{p} \in \mathbb{R}^{L \times C}$ are then added to the patch embeddings to incorporate spatial information, resulting in the final patch tokens $\textbf{x}_{\text{token}} = \textbf{x}_e + \textbf{x}_p$.

\subsection{Transformer Block = Attention + Feed-forward Network}
Transformer Block aggregates information carried by tokens globally to refine the tokens into higher semantic representations. Tokens will better represent the concepts for the patches as they pass through more transformer blocks. A transformation block consists of an Attention layer and a Feed-forward Network (FFN) with residual connections.  

\subsubsection{Multi-head Attention}
\begin{figure}
    \centering
    \includegraphics[width=0.25\linewidth]{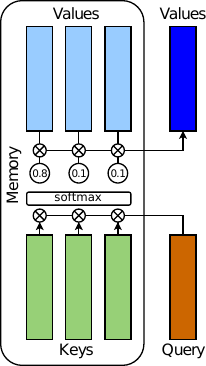}
    \caption{Overview of Attention mechanism.}
    \label{fig:attn}
\end{figure}
The Attention mechanism in Vision Transformers allows the model to weigh the importance of different patches relative to each other. This is achieved through multi-head attention, where the model can attend to different positions in the image simultaneously. The mechanism involves generating query ($\mathbf{Q}$), key ($\mathbf{K}$), and value ($\mathbf{V}$) vectors from the input tokens. \cref{fig:attn} illustrates the basic attention mechanism. The attention weights are computed using the scaled dot-product attention formula:

\begin{equation}\label{eq:attn}
    \text{Attention}(\mathbf{Q}, \mathbf{K}, \mathbf{V}) = \text{softmax}\left(\frac{\mathbf{Q}\mathbf{K}^T}{\sqrt{d_k}}\right)\mathbf{V},
\end{equation}
where $d_k$ is the dimensionality of the key vectors. The softmax function ensures that the attention weights sum to 1, enabling the model to focus on relevant patches for each token. Multi-head attention extends this by computing attention in parallel over multiple heads, allowing the model to capture diverse relationships:

\begin{equation}\label{eq:mha}
\text{MultiHead}(\mathbf{Q}, \mathbf{K}, \mathbf{V}) = \text{Concat}(\text{head}_1, \dots, \text{head}_h)\mathbf{W}^O,
\end{equation}
where each head is computed as $\text{head}_i = \text{Attention}(\mathbf{Q}\mathbf{W}_i^Q, \mathbf{K}\mathbf{W}_i^K, \mathbf{V}\mathbf{W}_i^V)$, and $\mathbf{W}_i^Q$, $\mathbf{W}_i^K$, $\mathbf{W}_i^V$, and $\mathbf{W}^O$ are learnable projection matrices. $\mathbf{W}^O$ is able to fuse the learned information from these heads.

\subsubsection{Feed-forward Network}
The Feed-forward Network (FFN) is a standard MLP applied independently to each patch token. It helps in modeling complex relationships and transformations of the features extracted by the Attention layer. The FFN typically consists of two linear layers with an activation function, ReLU or GELU. The FFN can be expressed as:

\begin{equation}\label{eq:ffn}
\text{FFN}(\mathbf{x}) = \mathbf{W}_2^T \sigma(\mathbf{x}\mathbf{W}_1 + \mathbf{b}_1) + \mathbf{b}_2,
\end{equation}
where $\mathbf{W}_1$, $\mathbf{W}_2$ are learnable weight matrices, and $\mathbf{b}_1$, $\mathbf{b}_2$ are bias terms, $\sigma(\cdot)$ denotes an activation function, e.g., ReLU. Like the Attention layer, it uses residual connections and layer normalization to facilitate better training and convergence.

The FFN can be treated as a memory-based attention module~\citep{geva2020transformer} in which the first layer is a set of keys and the second layer stores the corresponding keys.

\subsection{Task-related Head}

The Projection Head is a critical component in Vision Transformers (ViTs) for adapting the model to different downstream tasks, such as classification~\citep{touvron2021training} and dense prediction~\citep{ranftl2021vision}. In this report, we adopt classification as an example to demonstrate how to utilize the extracted features from tokens for downstream tasks.

A [CLS] token is inserted alongside the patch tokens before being processed by the Transformer blocks. The [CLS] token serves as a global representation of the entire image, enabling the model to focus on global information rather than local information tied to specific positions. After passing through the Transformer blocks, the final representation of the [CLS] token (or a global pooling of all tokens) is fed into a projection head. This projection head maps the high-dimensional features to a task-specific output space. 

The projection head is typically implemented as a simple multi-layer perceptron (MLP). For a classification task with $C$ classes, the projection head can be formulated as:
\begin{equation}
    \mathbf{p} = \text{MLP}(\mathbf{h}_{\text{[CLS]}}) = \mathbf{W}_2^T \sigma(\mathbf{W}_1  \mathbf{h}_{\text{[CLS]}} + \mathbf{b}_1) + \mathbf{b}_2,
\end{equation}
where $\mathbf{h}_{\text{[CLS]}} \in \mathbb{R}^C$ is the final representation of the [CLS] token, $\mathbf{W}_1 \in \mathbb{R}^{D \times C}$ and $\mathbf{W}_2 \in \mathbb{R}^{C \times D}$ are learnable weight matrices, $\mathbf{b}_1 \in \mathbb{R}^D$ and $\mathbf{b}_2 \in \mathbb{R}^C$ are bias terms, and $\text{ReLU}$ is the activation function. The output $\mathbf{p} \in \mathbb{R}^C$ represents the logits for each class.

The model is trained using the cross-entropy loss, which measures the discrepancy between the predicted class probabilities and the ground truth labels. The cross-entropy loss is defined as:
\begin{equation}
    \mathcal{L} = -\sum_{i=1}^C \mathbf{p}_i \log(\text{softmax}(\mathbf{y})_i),
\end{equation}
where $\mathbf{y} \in \{0, 1\}^C$ is the one-hot encoded ground truth label, and $\text{softmax}(\mathbf{p})_i$ is the predicted probability for class $i$.

\subsection{Modern Techniques}

The computational cost of the standard attention mechanism has a complexity of \(O(L^2·d + n·d^2)\), where \(L\) is the number of tokens. This quadratic complexity can become a bottleneck, especially when processing high-resolution images or long sequences. To address this, many modern techniques have been proposed to improve the efficiency of attention mechanisms. These techniques include sparse attention~\citep{child2019generating}, linear attention~\citep{katharopoulos2020transformers}, and low-rank compression~\citep{liu2024deepseek}. In this report, we focus on a mechanism called \textbf{multi-latent attention}(MLA), which compresses matrices to reduce KV cache while maintaining the ability to capture global dependencies.

\paragraph{Multi-latent Attention.}

\begin{figure}
    \centering
    \includegraphics[width=0.9\linewidth]{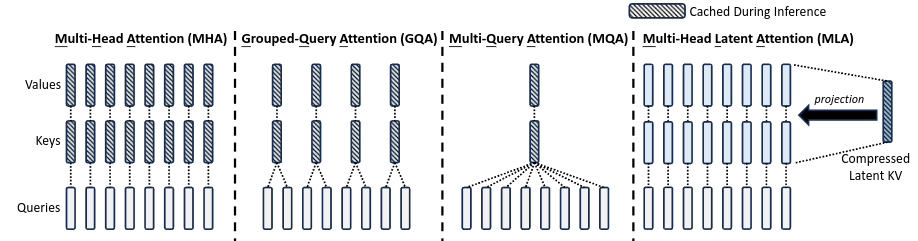}
    \caption{Illustration of Multi-Head Attention (MHA), Grouped-Query Attention (GQA), Multi-Query Attention (MQA), and Multi-head Latent Attention (MLA) from~\citep{liu2024deepseek}. }
    \label{fig:mla}
\end{figure}

The core of MLA is the low-rank compression to reduce active memory by a down-projection matrix $W^D$ and an up-projection matrix $W^U$. For instance, we can perform low-rank compression for the queries:
\begin{equation}
    \begin{split}
        c = W^D x,\\
        q  =W^U c,
    \end{split}
\end{equation}
where $c \in \mathbb{R}^{d_c}$ is the compressed latent vector; $d_c$ denotes the compression dimension, which is far smaller than the original dimension.

\section{Experiment}

\subsection{Tiny Vision Transformer}

\paragraph{Baseline.}
The baseline code is based on \cite{gani2022train}. A variety of data augmentation methods are applied, including CutMix~\citep{yun2019cutmix}, MixUp~\citep{zhang2017mixup}, AutoAugment (aa)~\citep{hoffer2020augment}, and Repeated Augment~\citep{cubuk2019autoaugment,hoffer2020augment}. Additionally, regularization methods such as label smoothing~\citep{Szegedy_2016_CVPR}, stochastic depth~\citep{huang2016deep}, and random erasing~\citep{zhong2020random} are employed. All models are trained for 100 epochs with a batch size of 256 on a single NVIDIA 3090 GPU paired with an AMD EPYC 7532 32-Core Processor. The Adam optimizer is used with a cosine learning rate schedule, starting at 0.002 and decayed using a cosine strategy with a 10-epoch warmup. The weight decay rate is set to 5e-2. The Tiny Vision Transformer consists of 12 attention heads, 9 transformer blocks, and 64 patches for images (each patch size is 4x4). The experimental results are summarized in \cref{tab:exps}.

\begin{figure}
    \centering
    \includegraphics[width=1\linewidth]{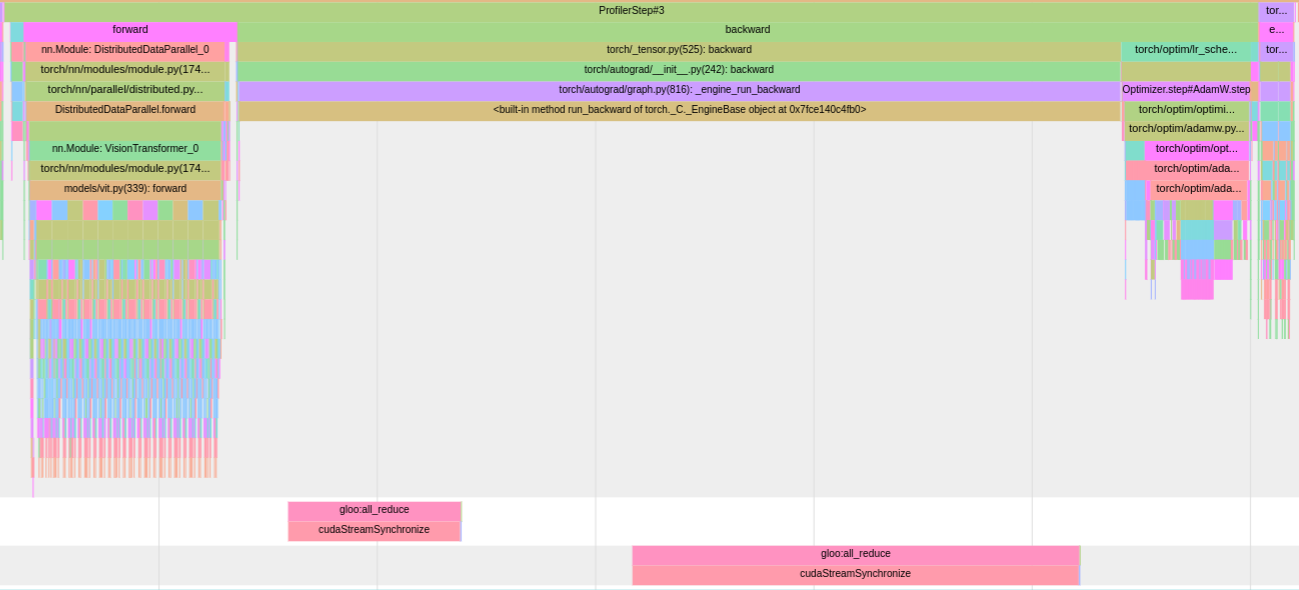}
    \caption{Tracing one training step. One step (120.7 ms) involves forward (25.2 ms), backward (79.3 ms), optimization (12.7 ms), and other processes.}
    \label{fig:profile}
\end{figure}

\paragraph{Profiler.}
To identify the training bottleneck, we employ the PyTorch Profiler~\citep{pytorchPyTorchProfiler}. \cref{fig:profile} visualizes the time span of each procedure using tracing\footnote{\url{edge://tracing/} in the Edge browser.}. The backward process consumes the majority of the training time, indicating that the bottleneck lies in the high computational cost of handling large feature maps during both forward and backward passes. \cref{tab:benchmark_tinyvit} compares the forward time across different batch sizes.

\begin{table}[ht]
    \centering
    \resizebox{0.8\textwidth}{!}{%
    \begin{tabular}{c|cccccc}
        \toprule
        \textbf{Batch Size (bs)} & \textbf{32} & \textbf{64} & \textbf{128} & \textbf{256} & \textbf{512} & \textbf{1024} \\ 
        \midrule
        \textbf{Throughput (images/s)} & 3162.09 & 6521.61 & 9449.16 & 9721.13 & 10368.24 & 10607.06 \\ 
        \textbf{Memory Usage (GB)} & 0.31 & 0.62 & 1.25 & 2.50 & 5.00 & 9.98 \\ 
        \bottomrule
    \end{tabular}%
    }
    \caption{Benchmark of Tiny ViT on an NVIDIA RTX 3090 GPU.}
    \label{tab:benchmark_tinyvit}
\end{table}

\begin{table}[ht]
    \centering
    \resizebox{\textwidth}{!}{
    \begin{tabular}{lcccccccccccccc}
        \toprule
        & \textbf{Baseline} & \multicolumn{3}{c}{\textbf{Augmentation}} & \multicolumn{2}{c}{\textbf{Patch}} & \multicolumn{5}{c}{\textbf{MLA}} & \multicolumn{3}{c}{\textbf{DDP}} \\
        \cmidrule(r){3-5} \cmidrule(r){6-7} \cmidrule(r){8-12} \cmidrule(r){13-15}
        &          & -aa   & -mixup & -cutmix & sin   & Witen   & qkv   & kv   & qk   & q    & k    & bs256 &       bs1024 &\\
        \midrule
        \textbf{Runtime} & 4658     & 4684  & 4656   & 4651   & 4687  & 4676  & 4805 & 4756 & 4785 & 4716 & 4711 & 3582  &       1371 & 1545\\
        \textbf{Val/Acc} & 93.65    & 92.14 & 93.68  & 92.77  & 93.43 & 93.28 & 91.8 & 92.43 & 92.42 & 93.32 & 92.5 & 93.44 &    92.09& 92.38\\
        \bottomrule
    \end{tabular}}
    \caption{Summary of our experiments. `aa' denotes AutoAugment. Data augmentation is critical except for Mixup. Learnable positional embeddings demonstrate the best performance. Compressing queries during training exhibits the least performance drop. Increasing batch size reduces training time when Distributed Data Parallel (DDP) is applied, but at the cost of diminished performance. Lion~\citep{chen2023lion} is more effective than AdamW~\citep{loshchilov2019adamw} for optimizing the Tiny ViT.}
    \label{tab:exps}
\end{table}

\subsection{Ablation Study}

\paragraph{Augmentation.}  
The experiments demonstrate the significance of data augmentation methods in improving validation accuracy. Removing AutoAugment (`-aa') results in a noticeable drop in accuracy (92.14 vs. 93.65), highlighting its importance. Similarly, disabling CutMix (`-cutmix') leads to a decline (92.77 vs. 93.65). Interestingly, removing MixUp (`-mixup`) marginally improves accuracy (93.68), suggesting it may not be as beneficial in this setup. These results underline the importance of selecting appropriate augmentations tailored to the dataset and architecture.

\begin{figure}
    \centering
    \includegraphics[width=0.5\linewidth]{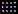}
    \caption{Whitening patterns in patch embedding.}
    \label{fig:whitening}
\end{figure}

\paragraph{Patch Token Initialization.}  
We utilize learnable positional embeddings by default. The positional embeddings can be replaced by sinusoidal positional embeddings (`sin'), which are non-learnable and fixed during training. Sinusoidal embeddings achieve slightly worse validation accuracy (93.43 vs. 93.65). We follow \citeauthor{ysam-code_cifar} to initialize the patch embedding using whitening initialization (`Witen') instead of random initialization. Whitening initialization identifies patterns in the dataset, as illustrated in \cref{fig:whitening}, to speed up the training process. However, this technique, while effective in CNNs, does not translate well to ViTs.

\paragraph{Distributed Data Parallel (DDP).}  
We conduct distributed data parallel training with 4 NVIDIA 3090 GPUs. Experiments with DDP show that increasing the batch size significantly reduces training time (e.g., from 3582s to 1371s with `bs1024'). However, larger batch sizes lead to diminished validation accuracy (92.09 vs. 93.65 for `bs256'). This trade-off emphasizes the need to balance computational efficiency with model performance when scaling training.

\begin{figure}
    \centering
    \includegraphics[width=0.7\linewidth]{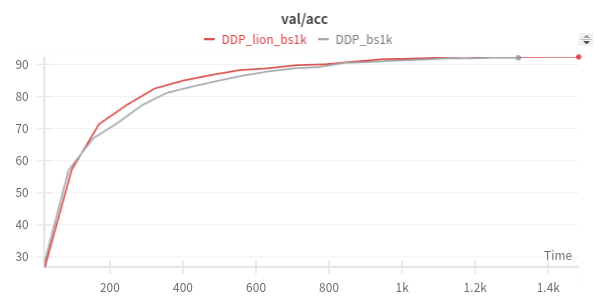}
    \caption{Accuracy over training time with the Lion optimizer. Although Lion consumes more time per step, it requires fewer iterations to find a good solution.}
    \label{fig:opt}
\end{figure}

\paragraph{Optimizer.}  
The Lion optimizer~\citep{chen2023lion} is compared against AdamW~\citep{loshchilov2019adamw} for optimizing the Tiny ViT. \cref{fig:opt} compares Lion to Adam in terms of finding a good solution. Lion achieves higher accuracy (92.38), demonstrating its superior capability for this architecture.

\subsection{Low-dimensional Patch Tokens}

\paragraph{Low-Rank Compression.}  
Different configurations of Multi-Head Latent Attention (MLA)~\citep{liu2024deepseek} were evaluated, including modifications to the queries (`q'), keys (`k'), or values (`v'). Compressing all three (`qkv') incurs a noticeable performance drop (91.8 vs. 93.65), while modifications targeting specific combinations (e.g., `kv', `qk', `q', and `k') yield better results. In particular, compressing `q' results in only a marginal drop in precision, consistent with the findings in \cite{liu2024deepseek}. These results suggest the potential for low-rank compression in ViTs to reduce activation memory.

\paragraph{Multi-Class Token.}
We investigate potential solutions to improve the efficiency of ViTs. One natural approach is to reduce the dimensionality of tokens. However, lower-dimensional tokens imply reduced capacity for visual information, as observed in \cref{tab:tokens}, where reducing the dimension to 96 results in a 3.24\% drop in accuracy. We argue that patch tokens, which are numerous (64 vs. 1), have a higher capacity than the CLS token. Even though patch tokens can represent the image effectively, these semantic representations are challenging to express through a narrow CLS token. To address this, we introduce additional CLS tokens to capture global representations. These CLS tokens are concatenated and fed to the classifier. Our simple multi-class token (MCLS) strategy effectively boosts the capacity of global representation. \cref{tab:tokens} validates our hypothesis, showing that using 2 CLS tokens improves accuracy from 90.41 to 93.95. Note that our MCLS approach differs significantly from Mctformer+~\citep{xu2024mctformer+}, where class information is distributed across multiple CLS tokens. In contrast, our proposal simply expands the dimensionality of the CLS token, with each CLS token representing a portion of the original CLS token.

\begin{table}[]
    \centering
\begin{tabular}{c|ccc}
\toprule
   dim  &  192 & 96 & 96\\
   num CLS & 1 &   1   & 2 \\
   
   \midrule
  acc & 93.65 & 90.41&  93.95 \\
  
  \bottomrule
  \end{tabular}
\caption{Multi-class token experiments.}
\label{tab:tokens}
\end{table}

{
    \small
    \bibliographystyle{ieeenat_fullname}
    \bibliography{main}

\begin{thebibliography}{28}
\providecommand{\natexlab}[1]{#1}
\providecommand{\url}[1]{\texttt{#1}}
\expandafter\ifx\csname urlstyle\endcsname\relax
  \providecommand{\doi}[1]{doi: #1}\else
  \providecommand{\doi}{doi: \begingroup \urlstyle{rm}\Url}\fi

\bibitem[Atito et~al.(2021)Atito, Awais, and Kittler]{atito2021sit}
Sara Atito, Muhammad Awais, and Josef Kittler.
\newblock Sit: Self-supervised vision transformer.
\newblock \emph{arXiv preprint arXiv:2104.03602}, 2021.

\bibitem[Chen et~al.(2023)Chen, Liang, Huang, Real, Wang, Liu, Pham, Dong, Luong, Hsieh, Lu, and Le]{chen2023lion}
Xiangning Chen, Chen Liang, Da Huang, Esteban Real, Kaiyuan Wang, Yao Liu, Hieu Pham, Xuanyi Dong, Thang Luong, Cho-Jui Hsieh, Yifeng Lu, and Quoc~V. Le.
\newblock Symbolic discovery of optimization algorithms, 2023.

\bibitem[Child et~al.(2019)Child, Gray, Radford, and Sutskever]{child2019generating}
Rewon Child, Scott Gray, Alec Radford, and Ilya Sutskever.
\newblock Generating long sequences with sparse transformers.
\newblock \emph{arXiv preprint arXiv:1904.10509}, 2019.

\bibitem[Cubuk et~al.(2019)Cubuk, Zoph, Mane, Vasudevan, and Le]{cubuk2019autoaugment}
Ekin~D Cubuk, Barret Zoph, Dandelion Mane, Vijay Vasudevan, and Quoc~V Le.
\newblock Autoaugment: Learning augmentation strategies from data.
\newblock In \emph{Proceedings of the IEEE/CVF conference on computer vision and pattern recognition}, pages 113--123, 2019.

\bibitem[Cydia2018(2021)]{Cydia2018}
Cydia2018, 2021.
\newblock https://github.com/Cydia2018/ViT-cifar10-pruning.

\bibitem[Dosovitskiy(2020)]{dosovitskiy2020image}
Alexey Dosovitskiy.
\newblock An image is worth 16x16 words: Transformers for image recognition at scale.
\newblock \emph{arXiv preprint arXiv:2010.11929}, 2020.

\bibitem[Gani et~al.(2022)Gani, Naseer, and Yaqub]{gani2022train}
Hanan Gani, Muzammal Naseer, and Mohammad Yaqub.
\newblock How to train vision transformer on small-scale datasets?
\newblock \emph{arXiv preprint arXiv:2210.07240}, 2022.

\bibitem[Geva et~al.(2020)Geva, Schuster, Berant, and Levy]{geva2020transformer}
Mor Geva, Roei Schuster, Jonathan Berant, and Omer Levy.
\newblock Transformer feed-forward layers are key-value memories.
\newblock \emph{arXiv preprint arXiv:2012.14913}, 2020.

\bibitem[Hoffer et~al.(2020)Hoffer, Ben-Nun, Hubara, Giladi, Hoefler, and Soudry]{hoffer2020augment}
Elad Hoffer, Tal Ben-Nun, Itay Hubara, Niv Giladi, Torsten Hoefler, and Daniel Soudry.
\newblock Augment your batch: Improving generalization through instance repetition.
\newblock In \emph{Proceedings of the IEEE/CVF Conference on Computer Vision and Pattern Recognition}, pages 8129--8138, 2020.

\bibitem[Huang et~al.(2016)Huang, Sun, Liu, Sedra, and Weinberger]{huang2016deep}
Gao Huang, Yu Sun, Zhuang Liu, Daniel Sedra, and Kilian~Q Weinberger.
\newblock Deep networks with stochastic depth.
\newblock In \emph{Computer Vision--ECCV 2016: 14th European Conference, Amsterdam, The Netherlands, October 11--14, 2016, Proceedings, Part IV 14}, pages 646--661. Springer, 2016.

\bibitem[Jordan(2024)]{jordan202494}
Keller Jordan.
\newblock 94\% on cifar-10 in 3.29 seconds on a single gpu.
\newblock \emph{arXiv preprint arXiv:2404.00498}, 2024.

\bibitem[Katharopoulos et~al.(2020)Katharopoulos, Vyas, Pappas, and Fleuret]{katharopoulos2020transformers}
Angelos Katharopoulos, Apoorv Vyas, Nikolaos Pappas, and Fran{\c{c}}ois Fleuret.
\newblock Transformers are rnns: Fast autoregressive transformers with linear attention.
\newblock In \emph{International conference on machine learning}, pages 5156--5165. PMLR, 2020.

\bibitem[Krizhevsky et~al.(2009)Krizhevsky, Hinton, et~al.]{krizhevsky2009learning}
Alex Krizhevsky, Geoffrey Hinton, et~al.
\newblock Learning multiple layers of features from tiny images.
\newblock 2009.

\bibitem[Liu et~al.(2024)Liu, Feng, Wang, Wang, Liu, Zhao, Dengr, Ruan, Dai, Guo, et~al.]{liu2024deepseek}
Aixin Liu, Bei Feng, Bin Wang, Bingxuan Wang, Bo Liu, Chenggang Zhao, Chengqi Dengr, Chong Ruan, Damai Dai, Daya Guo, et~al.
\newblock Deepseek-v2: A strong, economical, and efficient mixture-of-experts language model.
\newblock \emph{arXiv preprint arXiv:2405.04434}, 2024.

\bibitem[Liu et~al.(2021)Liu, Sangineto, Bi, Sebe, Lepri, and Nadai]{liu2021efficient}
Yahui Liu, Enver Sangineto, Wei Bi, Nicu Sebe, Bruno Lepri, and Marco Nadai.
\newblock Efficient training of visual transformers with small datasets.
\newblock \emph{Advances in Neural Information Processing Systems}, 34:\penalty0 23818--23830, 2021.

\bibitem[Loshchilov and Hutter(2019)]{loshchilov2019adamw}
Ilya Loshchilov and Frank Hutter.
\newblock Decoupled weight decay regularization, 2019.

\bibitem[Ranftl et~al.(2021)Ranftl, Bochkovskiy, and Koltun]{ranftl2021vision}
Ren{\'e} Ranftl, Alexey Bochkovskiy, and Vladlen Koltun.
\newblock Vision transformers for dense prediction.
\newblock In \emph{Proceedings of the IEEE/CVF international conference on computer vision}, pages 12179--12188, 2021.

\bibitem[Shao and Bi(2022)]{shao2022transformers}
Ran Shao and Xiao-Jun Bi.
\newblock Transformers meet small datasets.
\newblock \emph{IEEE Access}, 10:\penalty0 118454--118464, 2022.

\bibitem[Szegedy et~al.(2016)Szegedy, Vanhoucke, Ioffe, Shlens, and Wojna]{Szegedy_2016_CVPR}
Christian Szegedy, Vincent Vanhoucke, Sergey Ioffe, Jon Shlens, and Zbigniew Wojna.
\newblock Rethinking the inception architecture for computer vision.
\newblock In \emph{Proceedings of the IEEE Conference on Computer Vision and Pattern Recognition (CVPR)}, 2016.

\bibitem[Torch()]{pytorchPyTorchProfiler}
Torch.
\newblock {P}y{T}orch {P}rofiler.
\newblock \url{https://pytorch.org/tutorials/recipes/recipes/profiler_recipe.html}.
\newblock [Accessed 06-01-2025].

\bibitem[Touvron et~al.(2021)Touvron, Cord, Douze, Massa, Sablayrolles, and J{\'e}gou]{touvron2021training}
Hugo Touvron, Matthieu Cord, Matthijs Douze, Francisco Massa, Alexandre Sablayrolles, and Herv{\'e} J{\'e}gou.
\newblock Training data-efficient image transformers \& distillation through attention.
\newblock In \emph{International conference on machine learning}, pages 10347--10357. PMLR, 2021.

\bibitem[Vaswani(2017)]{vaswani2017attention}
A Vaswani.
\newblock Attention is all you need.
\newblock \emph{Advances in Neural Information Processing Systems}, 2017.

\bibitem[Xu et~al.(2024)Xu, Bennamoun, Boussaid, Laga, Ouyang, and Xu]{xu2024mctformer+}
Lian Xu, Mohammed Bennamoun, Farid Boussaid, Hamid Laga, Wanli Ouyang, and Dan Xu.
\newblock Mctformer+: Multi-class token transformer for weakly supervised semantic segmentation.
\newblock \emph{IEEE transactions on pattern analysis and machine intelligence}, 2024.

\bibitem[ysam code(2023)]{ysam-code_cifar}
ysam code, 2023.
\newblock https://github.com/tysam-code/hlb-CIFAR10.

\bibitem[Yun et~al.(2019)Yun, Han, Oh, Chun, Choe, and Yoo]{yun2019cutmix}
Sangdoo Yun, Dongyoon Han, Seong~Joon Oh, Sanghyuk Chun, Junsuk Choe, and Youngjoon Yoo.
\newblock Cutmix: Regularization strategy to train strong classifiers with localizable features.
\newblock In \emph{Proceedings of the IEEE/CVF international conference on computer vision}, pages 6023--6032, 2019.

\bibitem[Zhang(2017)]{zhang2017mixup}
Hongyi Zhang.
\newblock mixup: Beyond empirical risk minimization.
\newblock \emph{arXiv preprint arXiv:1710.09412}, 2017.

\bibitem[Zhang et~al.(2025)Zhang, Xu, Luo, and Wang]{ZHANGconvvit}
Tianxiao Zhang, Wenju Xu, Bo Luo, and Guanghui Wang.
\newblock Depth-wise convolutions in vision transformers for efficient training on small datasets.
\newblock \emph{Neurocomputing}, 617:\penalty0 128998, 2025.

\bibitem[Zhong et~al.(2020)Zhong, Zheng, Kang, Li, and Yang]{zhong2020random}
Zhun Zhong, Liang Zheng, Guoliang Kang, Shaozi Li, and Yi Yang.
\newblock Random erasing data augmentation.
\newblock In \emph{Proceedings of the AAAI conference on artificial intelligence}, pages 13001--13008, 2020.

\end{thebibliography}
}


\end{document}